\pdfoutput=1

\documentclass[11pt]{article}
\usepackage[table]{xcolor}

\usepackage[]{acl}

\usepackage{times}
\usepackage{latexsym}
\usepackage{multirow}
\usepackage{multicol}

\usepackage{tikz}
\newcommand{\tikzxmark}{%
\tikz[scale=0.23] {
    \draw[line width=0.7,line cap=round] (0,0) to [bend left=6] (1,1);
    \draw[line width=0.7,line cap=round] (0.2,0.95) to [bend right=3] (0.8,0.05);
}}
\newcommand{\tikzcmark}{%
\tikz[scale=0.23] {
    \draw[line width=0.7,line cap=round] (0.25,0) to [bend left=10] (1,1);
    \draw[line width=0.8,line cap=round] (0,0.35) to [bend right=1] (0.23,0);
}}

\usepackage[T1]{fontenc}

\usepackage[utf8]{inputenc}

\usepackage{microtype}

\title{Improving Health Question Answering with \\ Reliable and Time-Aware Evidence Retrieval}

\author{Juraj Vladika \and Florian Matthes \\
  Department of Computer Science \\ Technical University of Munich \\ Garching, Germany \\
  \texttt{ \{juraj.vladika, matthes\}@tum.de} \\}

\begin{document}
\maketitle
\begin{abstract}
In today's digital world, seeking answers to health questions on the Internet is a common practice. However, existing question answering (QA) systems often rely on using pre-selected and annotated evidence documents, thus making them inadequate for addressing novel questions. Our study focuses on the open-domain QA setting, where the key challenge is to first uncover relevant evidence in large knowledge bases. By utilizing the common retrieve-then-read QA pipeline and PubMed as a trustworthy collection of medical research documents, we answer health questions from three diverse datasets. We modify different retrieval settings to observe their influence on the QA pipeline's performance, including the number of retrieved documents, sentence selection process, the publication year of articles, and their number of citations. Our results reveal that cutting down on the amount of retrieved documents and favoring more recent and highly cited documents can improve the final macro F1 score up to 10\%. We discuss the results, highlight interesting examples, and outline challenges for future research, like managing evidence disagreement and crafting user-friendly explanations.
\end{abstract}

\section{Introduction}

In the digital era, using the Internet to search for health information has become a prevalent behavior \cite{healthcare9121740}. Users turn to seek health advice online due to its ease of access, wide coverage of information, convenience of searching, interactivity, and anonymity \cite{neely2021health}. Health information sought online includes anything regarding the symptoms and treatments of different diseases. In general, health information seeking can lead to enhanced patient involvement in medical decision-making, improved communication with care providers, and improved quality of life \cite{doi:10.1177/0033354919874074}. Nevertheless, finding trustworthy and relevant evidence in abundant online content remains an open challenge \cite{battineni2020factors}. Especially in the medical field, clinical recommendations can change with time, so finding the latest evidence is essential for reliable answers.

\begin{figure}[t]
  \centering
  \includegraphics[width=0.99\linewidth]{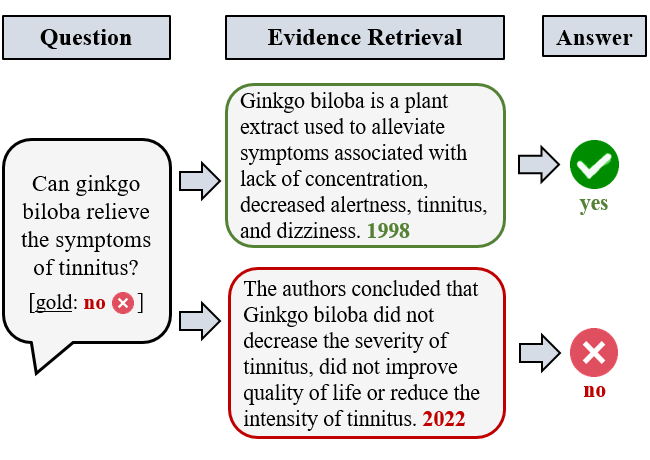}
  \caption{The question-answering system used in our study and an example question with two retrieved evidence documents and final predictions. This example shows how retrieving an outdated study caused an incorrect prediction (top), while a more recent study resulted with an accurate answer (bottom).}
  \label{fig:pipeline}
\end{figure}

Interacting with online search engines and conversational systems is done with \textit{question answering (QA)}. Technical solutions based on Machine Learning (ML) and Natural Language Processing (NLP) aim to automate this task and provide users with reliable answers to their inquiries. The purpose of QA systems is multi-fold: it helps scientists verify their research hypothesis by finding related studies, it allows lay users to find answers to their everyday health concerns, and enables factuality assessment of generative language models by fact-checking their responses over trusted evidence \cite{jin2021disease, vladika-matthes-2023-scientific}.

While QA can work with answering questions over a provided document, we are focusing on the more realistic and challenging problem of \textit{open-domain question answering}, where extensive collections of documents with diverse topics have to be quried to find the relevant evidence \cite{chen2020open}. The open-domain QA systems usually consist of two main components, a retriever and a reader \cite{zhu2021retrieving}. The retriever's task is finding the relevant documents that will serve as the main source of evidence for answering the question. The reader (QA module) performs the reasoning process between the question and found evidence, and produces the final answer. While both components are essential for the system, we posit that the retrieval is a more challenging part, considering that the QA module receives the input from it and the quality of the final answer depends on the retrieved documents \cite{sauchuk2022role}. 

Retrieving credible evidence documents relevant to the query ensures the final output's quality. This is true for both the retrieval-powered text classification tasks and recently popular retrieval-augmented generation (RAG) approaches \cite{cuconasu2024power}. While much progress has been made in open-domain QA, addressing the challenges in retrieval settings still needs to be explored. These include assessing the quantity of documents needed to be retrieved for a reliable answer, the amount of evidence passages extracted from them, and the quality of the documents themselves, such as their recency and strength of findings. Figure \ref{fig:pipeline} shows an example of a health question answered with two different retrieved documents -- the more recent one has more up-to-date knowledge and findings.

To bridge this research gap, in this study, we perform an array of experiments to test the predictive performance of an open-domain QA system with different evidence retrieval strategies. We use three diverse datasets of biomedical and health questions, which contain discreet labels like \textit{yes} and \textit{no} as their final answers, and use their gold labels as ground truth. We use the large collection of 20 million biomedical research abstracts from PubMed as the knowledge base for evidence retrieval. By keeping the reader (answering module) fixed, we only vary the different retrieval aspects and measure the change in the QA performance as measured by classification metrics precision, recall, and F1. The settings we explore include the number of documents retrieved and sentences extracted from them, the articles' publication year, and their number of citations. Our findings demonstrate that the QA performance is improved by accounting for the amount and quality of the retrieved documents.

To summarize, our research contributions are:
\begin{itemize}
    \item We evaluate the performance of an open-domain QA pipeline for health questions, using biomedical research papers as evidence source, concerning the different number of documents retrieved and sentences extracted from them.
    \item Additionally, we evaluate the influence of the evidence quality parameters like year of publication and number of citations on the final predictive performance of the QA system, showing that time-aware evidence retrieval leads to improved performance.
    \item Finally, we take a deeper look into the results and provide insights with a qualitative analysis. We report on the problem of evidence disagreement and provide future directions on developing better health question answering systems to be deployed in the future.
\end{itemize}

We make our code and datasets publicly available in a GitHub repository.\footnote{\url{https://github.com/jvladika/Improving-Health-QA}}

\section{Related work}
In this section, we outline the work related to our study. 

\subsection{Biomedical Question Answering}
Question Answering (QA) is a rapidly evolving knowledge-intensive NLP task, with over 80 QA datasets released in last two years \cite{rogers2023qa}. Based on the availability of evidence for the question, it can be analyzed in a closed-domain or an open-domain setting. In closed-domain QA, the evidence comes from an already provided source document. This setting is also called Machine Reading Comprehension (MRC) since the goal is to build models that can comprehend from the given text how to answer the posed question \cite{baradaran2022survey}. In open-domain QA, to which our work belongs, only the question and its final answer are known, and the QA system has to find appropriate evidence in a large document corpus or other type of collection \cite{chen-yih-2020-open}. 

Based on the topic of questions, our work is related to the research on \textit{science question answering}, aiming to answer questions related to natural sciences from resources like school curricula \cite{lu2022learn} or scholarly articles \cite{lee2023qasa}. More precisely, our work is part of  \textit{biomedical question answering} \cite{jin2022biomedical}. 

Biomedical QA can help biomedical researchers conduct their work by answering complex research questions \cite{jin-etal-2019-pubmedqa}, help clinical practitioners by answering clinical questions over health records \cite{vilares-gomez-rodriguez-2019-head}, or help consumers answer questions about their health concerns \cite{demner2020consumer}. The last category, also called \textit{health question answering}, is increasingly being adopted by consumers due to the rising popularity of conversational assistants \cite{budler2023review}. In our work, we cover both the datasets related to QA for biomedical research and consumer health. 


\subsection{Open-Domain Fact Verification and QA}
Considering that the datasets we use only contain questions with discrete (\textit{yes/no}) answers, our work is related to the task of automated fact verification (fact-checking). This task aims to verify the veracity of a factual claim based on credible evidence that supports it, refutes it, or does not provide enough information \cite{guo2022survey}. Recent years have seen a rise in fact-checking datasets focusing on scientific knowledge, in particular health and medicine \cite{vladika-matthes-2023-scientific}. 

While fact verification literature often works in a closed-domain setting with evidence documents provided, some recent work also explores the open-domain setting.  \citet{wadden-etal-2022-scifact} observe significant performance drops in F1 scores of final verdict predictions when increasing the evidence corpus from a few thousand to half a million documents. \citet{pugachev2023consumer} analyze how well consumer-health questions can be answered with built-in search engines of PubMed and Wikipedia. Expanding the scope even more, \citet{vladika-matthes-2024-comparing} compare the performance of semantic search and BM25 over PubMed and Wikipedia, as well as Google search, for verifying biomedical and health questions.

Some studies have analyzed the influence of time and quantity in evidence retrieval on downstream tasks. \citet{allein2021time} trained time-aware evidence ranking models for time-sensitive news claims and show performance improvement. Likewise, \citet{schlichtkrull2024averitec} constructed a dataset where evidence for given claims only appeared after the claim itself, thus eliminating temporal leaks. Regarding the document quantity, \citet{oh-thorne-2023-detrimental} analyze the influence of the number of retrieved evidence passages on the QA performance over two general QA datasets, showing that the performance often actually drops with the increasing number of retrieved snippets. 

To the best of our knowledge, our paper uses the largest document collection so far for open-domain health QA by indexing the entire PubMed corpus of more than 20 million articles. Likewise, it is also the first study to test the influence of the number of documents retrieved on the final QA performance instead of fixing it to a certain number, like the commonly found 5 \cite{thorne-etal-2018-fever} or 6 \cite{wadden-etal-2022-scifact}; as well as testing the influence of the different number of sentences retrieved. While there has been existing research on time-aware evidence ranking in fact verification for news claims, our work is first to explore the temporal aspect for biomedical questions, as well as other evidence quality aspects like the number of citations of retrieved publications. 

\section{Foundations}
In this section, we explain the foundations of the study, including the used datasets, the evidence corpus, and the structure of the used QA system.

\subsection{Datasets}
We chose three datasets of biomedical and health claims in English, built for different purposes. We only use the questions and final labels (answers) from the datasets in our experiments. While the datasets provide the gold evidence passages used to derive the answers, we do not utilize them since the idea of our open-domain QA setting is that the retriever component has to discover the relevant evidence in a large document corpus.

\textbf{HealthFC} \cite{vladika2023healthfc} is a question-answering and fact-checking dataset focusing on consumer health questions and common topics users search health advice online for. It includes diverse topics like dietary supplements, heart and lungs, reproductive health, cancer, and mental health. Medical practitioners manually answered and verified all the questions using the evidence from systematic reviews and clinical trials. There are 750 questions in total, out of which 205 are supported, 122 are refuted, and for 433 questions, there is not enough information (NEI) to answer. We use two variants of the dataset: \textbf{HealthFC-3}, which has all \textbf{750} claims and all three classes; and \textbf{HealthFC-2}, which only has \textbf{327} supported and refuted claims with two classes.

\textbf{BioASQ-7b} \cite{10.1007/978-3-030-43887-6_51} is a biomedical question-answering dataset constructed by biomedical researchers and designed to reflect the real information needs of biomedical experts. It is part of the ongoing challenge of the same name, focusing on biomedical semantic indexing and question answering. The evidence for answers comes from biomedical research publications, i.e., the same corpus of PubMed used in our study. Other than only exact answers, the BioASQ dataset also includes ideal answer summaries. The 7b version of the dataset we use has \textbf{745} claims, of which 614 are supported ("yes"), and 131 are refuted ("no").

\textbf{TREC-Health} \citep{pugachev2023consumer} is a dataset of 117 popular health questions originating from two TREC shared challenges. TREC is an ongoing series of workshops centering on challenges in accurate information retrieval \cite{voorhees2005trec}. The questions stem from two shared tasks: the TREC 2019 Decision Track \cite{Abualsaud2020OverviewOT} and the TREC 2021 Health Misinformation Track \cite{Clarke2021OverviewOT}, both focusing on challenges with incorrect search engine results for health (mis)information. Questions cover common consumer health concerns, similar to HealthFC, but the two datasets do not overlap. The dataset consists of \textbf{113} questions, of which 61 are supported ("yes") and 52 refuted ("no").

Table \ref{tab:datasets} gives an overview of the four datasets. With HealthFC and TREC-Health, we aim to target common health questions users would pose to a QA system, while BioASQ is selected in order to explore how do the QA results change for more complex, expert-geared biomedical questions.

\begin{table}[htpb]
\begin{tabular}{p{21mm}|p{17mm}|ccc}
\hline
\textbf{Dataset}  & \textbf{Domain}                & \tikzcmark & \tikzxmark & \textbf{?} \\
\hline
\textbf{HealthFC-3} & everyday health  & 202 & 125 & 433 \\ \hline
\textbf{HealthFC-2} & everyday health  & 202 & 125 & --- \\ \hline
\textbf{BioASQ-7b}   & biomedical research & 614 & 131  & --- \\ \hline
\textbf{TREC-Health}   & consumer health & 61 & 52  & --- \\ 
\hline
\end{tabular}

\caption{\label{tab:datasets} The four datasets used in the experiments, including their domain and label distribution. (\textbf{\tikzcmark} -- supported, \tikzxmark -- refuted, \textbf{?} -- not enough information)}
\end{table}

\begin{table*}[htpb]
\centering
\rowcolors{2}{gray!25}{white}
\begin{tabular}{r|ccc|ccc|ccc|ccc}

\rowcolor{gray!40}
\textbf{Top k} & \multicolumn{3}{c}{\textbf{HealthFC-3}} & \multicolumn{3}{c}{\textbf{HealthFC-2}} & 
\multicolumn{3}{c}{\textbf{TREC}} & \multicolumn{3}{c}{\textbf{BioASQ}}  \\
\cline{2-13}
\rowcolor{gray!40}
  \textbf{docs}    & \textbf{P}       & \textbf{R}       & \textbf{F1}  & \textbf{P}       & \textbf{R}       & \textbf{F1}       & \textbf{P}      & \textbf{R}      & \textbf{F1}     & \textbf{P}       & \textbf{R}      & \textbf{F1}       \\
\hline
\textbf{1}       &   49.2     &     44.1     &   \textbf{40.1}    &    63.3  &   \textbf{62.7}    &   \textbf{62.9}      &   61.7     &   60.9    &  \textbf{60.7}   &    \textbf{71.8}     &    57.4     &   58.8           \\
\textbf{5}     &      53.9     &     44.7   &    38.8     &     \textbf{67.6 }   &     57.8   &    55.2       &   63.8   &   58.7  &  55.8    &   65.5     &    \textbf{65.0 }   &    65.2            \\
\textbf{10}      &    \textbf{55.1}    &   \textbf{45.2}      &   39.5     &    66.4     &     56.5    &      53.1    &  \textbf{67.6}     &  \textbf{63.0}  &   57.2   &   67.7     &    63.9     &   \textbf{65.4}           \\
\textbf{15}     &     58.2    &    44.4      &     39.1    &    67.5     &    55.6     &     51.1     &   65.6     &   58.4     &    54.5    &    68.2     &   63.4     &    65.1     \\
\textbf{20}    &   49.8      &   43.7      &   38.5      &    63.8    &     54.9    &    50.5        &    66.9   &   58.3     &   53.7     &   67.5     &   63.0      &    64.5         \\
\textbf{50}     &     48.4    &  42.9       &  38.1      &   62.2     &    53.3   &   47.6    &    67.5    &   57.1   &   51.4   &    67.5     &   60.4     &   62.1  \\
\textbf{100}    &    47.5     &    44.3   &    37.6     &    56.3    &     44.6   &    44.6     &     64.5   &   56.3     &   50.8     &    66.3     &   59.6     &  61.2

\end{tabular}
\caption{\label{tab:topk} Results of final answer prediction over four datasets, with different values of \textit{top k} documents retrieved during the process and used for majority voting. All scores are macro averaged.}

\end{table*}

\subsection{Evidence Corpus}
We approach the QA task in the open-domain setting, meaning that evidence is unknown when the question is posed and must first be discovered in a vast evidence collection. Given that we work with medical and health-related questions, we chose a collection of biomedical research publications as the source of evidence.

Our evidence corpus originates from PubMed, a large and trustworthy knowledge base of biomedical research publications \cite{canese2013pubmed}. Considering that the full text of most of these publications is not freely accessible, we use only the abstracts, which are always available. This does not hinder the performance since medical abstracts often already include a verdict on their main research hypothesis. The US National Library of Medicine provides every year MEDLINE, a snapshot of currently available abstracts in PubMed that is updated once a year. We used the 2022 version found on the official website.\footnote{\url{https://www.nlm.nih.gov/databases/download/pubmed_medline.html}} While this yields 33.4M abstracts, we pre-processed the data by removing any non-English papers, papers with no abstracts, and papers with unfinished abstracts, which yields 20.6 million abstracts. 

\subsection{QA System}
The question-answering system used for our experiments is in the form of a pipeline, based on the pipeline system from \citet{vladika-matthes-2023-sebis}. This pipeline consists of two main parts: a \textit{retriever} and a \textit{reader}. The process of question answering is done sequentially, by first retrieving the evidence, performing reasoning over it, and finally producing a final answer. Our experiments focus on changing the different aspects of the retriever while keeping the reader completely fixed. This ensures that the experimental setup is consistent and that only one parameter is tested at a time.

In the retriever, given a question $q$ and a corpus of $n$ documents $D = \{d_1, d_2, ..., d_n\}$, the goal is to select the top $k$ most relevant documents $g_1, ..., g_k$ for the given query. The selection is done with a function $search(q, d)$, which compares the similarity of the question (query) and each document in the corpus. The documents in our corpus are abstracts of medical publications. While abstracts are shorter versions of full documents, they can still contain irrelevant sentences for producing the final verdict. 

In our first experiment, we use full documents $g_1, ..., g_k$ and the question $q$ to predict the final answer. In the second experiment, we select only the top $j$ most relevant sentences from the abstracts.
From $m$ candidate sentences $s_1, s_2, ..., s_m$ comprising the selected documents, top $j$ sentences are selected as evidence sentences $\vec{e} = e_1, e_2, ..., e_n$ with a function $select(q, s)$. These sentences are the most similar to the question $q$.

Finally, the answer is predicted from the given question $q$ and evidence $\vec{e}$, where the evidence is either the complete documents (the first round of experiments) or a set of sentences (the second round of experiments). The reader model produces the final verdict and is one of three classes $y(q, \vec{e}) \in \{\textit{Refuted} (0)$, $\textit{Supported} (1)$, $\textit{Not Enough Information} (2) \}$. While QA can be generative and elicit long answers, all datasets we use contain only the short \textit{yes/no/unknown} answers. This makes using the standard classification metrics precision, recall, and F1 possible. We model answer prediction as the related task of recognizing entailment or natural language inference (NLI) and use an NLI model for the prediction.

In all experiments, majority voting is used to determine the final verdict. For the dataset HealthFC-3 with three classes, this can be one of the three classes (0, 1, 2). For other datasets, the majority is taken only from predictions of 0 and 1 (in case of a tie, 0 is predicted). Majority voting is chosen for convenience, but it is not optimal as the information on prediction disagreement. Future work should explore how to model the disagreement better.

For the $search(q, d)$ function that selects \textit{top k} most relevant documents, we use BM25 as it was proven to be a strong baseline for retrieving documents for automated claim verification \cite{stammbach2023choice}. It also ensures higher precision at the cost of coverage, which aligns with our use case -- we want the retrieved documents to be relevant before being passed to the reader. We use a sentence embedding model and semantic search to select the sentences most similar to our query from abstracts. For $select(q, s)$, we select the model \textsc{Spiced} \cite{wright-etal-2022-modeling}, which is a recent sentence similarity model optimized for paraphrases of scientific claims. For the final answer prediction model (reader) $y(q, \vec{e})$, we choose the DeBERTa-v3 model \cite{he2021deberta} since it was shown to be a highly potent model for natural language understanding and reasoning tasks. We use the variant of the model optimized for NLI prediction \cite{laurer2024less}. We do not fine-tune the models on the datasets in our experiments because we want to simulate a realistic QA system that has to answer unseen questions.

\begin{table*}[htpb]
\centering
\rowcolors{2}{gray!25}{white}
\begin{tabular}{r|ccc|ccc|ccc|ccc}

\rowcolor{gray!40}
\textbf{Top j} & \multicolumn{3}{c}{\textbf{HealthFC-3}} & \multicolumn{3}{c}{\textbf{HealthFC-2}} & \multicolumn{3}{c}{\textbf{TREC}} & \multicolumn{3}{c}{\textbf{BioASQ}}  \\

\cline{2-13}
\rowcolor{gray!40}
 \textbf{sents}    & \textbf{P}       & \textbf{R}       & \textbf{F1}  & \textbf{P}       & \textbf{R}       & \textbf{F1}       & \textbf{P}      & \textbf{R}      & \textbf{F1}     & \textbf{P}       & \textbf{R}      & \textbf{F1}       \\
\hline
\textbf{1}  & 40.7 &  40.0 &  35.9  &  57.4 &   57.4  &  \textbf{57.4}    &    59.8       &    \textbf{59.3 }    &    59.1    &      56.9     &    62.0     &    52.6     \\
\textbf{3}  & 39.0 &  41.7 & 36.6 &  60.2    &  56.9  &  56.0  &    59.3       &      57.6   &    56.4    &      58.8     &     64.1    &    58.2        \\
\textbf{5}  &  52.2 & \textbf{ 44.6} &  39.8 & 59.8  &  56.3  & 54.9  &     59.7      &     57.3    &   \textbf{59.3}     &      58.5     &     62.8    &     58.6       \\
\textbf{10}  & 46.9 & 44.4 &  40.0 &   61.4  &  56.9  & 55.3  &    \textbf{63.1  }    &    58.8    &   56.4   &   50.5     &     62.5     &    60.2   \\
\textbf{15}  &  45.4 & 44.2 & 39.8  &  61.9  &  57.3  &   55.8  &    60.9    &   56.9      &    53.8    &      60.7     &     64.2    &     61.5       \\
\textbf{20}  &  \textbf{52.6 } &  44.5  & \textbf{40.6} &  \textbf{64.3}  & \textbf{ 58.2 }  &  56.8  &     61.5    &    58.0    &    55.7   & \textbf{ 60.9  }  &   \textbf{ 64.4}     &     \textbf{61.7 }
\end{tabular}

\caption{\label{tab:topj} Results of final answer prediction over four datasets, with different values of \textit{top j} sentences retrieved during the process and used for majority voting. All scores are macro averaged.}
\end{table*}

\section{Experiments}
We conduct three groups of experiments to test the influence of different retrieval parameters on the performance of our QA system.

\paragraph{Number of retrieved documents.} The first group of experiments consisted of keeping the QA pipeline consistent but increasing the number of retrieved documents (\textit{top k}) that are forwarded to the final QA module. The motivation behind this was to find the fine balance between covering enough different studies but not saturating the module with noise and irrelevant articles. Considering the increasing popularity of retrieval-enhanced systems such as retrieval-augmented generation (RAG) pipelines \cite{lewis2020retrieval}, retrieving only the relevant amount of documents or chunks is a significant challenge. We use BM25 as the retrieval technique because of its efficiency and its focus on enhancing precision instead of recall.

\paragraph{Number of retrieved sentences.} Instead of taking the whole documents and sending them with the question to the reader, the second group of experiments selected only the \textit{top j} most relevant sentences within all abstracts and used those as evidence. In this setup, we first retrieve the top 20 most similar abstracts with BM25. Afterward, all abstracts are split into sentences, which are embedded with the sentence-embedding model SPICED. After that, the \textit{top j} most similar sentences to the question \textit{q}, according to cosine similarity, are chosen from the pool of sentences (so multiple sentences can come from the same abstract). The QA module calculates the entailment probability between the question and each selected sentence, and finally, majority voting is performed.

\paragraph{Year of publication and number of citations.} The third and fourth group of experiments focused less on the technical parameters of the retrieval but more on the \textit{quality} of the discovered evidence. So far, not many studies have leveraged the metadata of retrieved evidence documents for enhancing medical and health-related question answering or fact verification. We use two metadata parameters that should intuitively have an influence on the quality of the performance -- \textit{year of publication} of the retrieved research publication and the \textit{number of citations} it has. The year was provided among the metadata that comes with PubMed, but getting the number of citations was more challenging, considering it is not foundin the MEDLINE dump. Therefore, we utilized the \textit{Semantic Scholar API} \cite{ammar-etal-2018-construction} by querying it with the PubMed ID (PMID) of the retrieved article and then calling the API to get the number of citations. Once we had both numbers, experiments consisted of filtering the pool of possible evidence documents by posing a restriction on the minimum year of publication and the minimum number of citations. Out of the top \textit{k} documents we retrieved, only those published after a certain year, or those with at least a certain number of citations, were selected as the final evidence documents. The rest of the workflow is the same as in the first experiment: the documents are passed to the QA module and the answer is predicted.

Given that we work with discrete answers and labels, the evaluation metrics we used are macro-averaged versions of classification metrics precision, recall, and F1 score. Macro averaging implies that the arithmetic mean of the metric for each individual class is taken (e.g., $F1_{macro} = \sum_{i} F1_{class \: i} / n $). For HealthFC-3, this is an average of three classes, while for the other datasets, it is an average of two classes. The motivation behind using the macro version is that it considers all classes equally important. We posit that in a deployed health QA system, users would be interested not only in detecting positive answers, but the system effectively discerning between both negative and positive answers to questions. This also follows the literature on automated fact-checking, which commonly uses macro-averaged scores \cite{bekoulis2021review}.

All the experiments were run on a single Nvidia V100 GPU with 16 GB of VRAM. The process of retrieving the top 100 most relevant documents for each dataset used one computation hour. The process of predicting the final answer with the top 100 most relevant documents also used one computation hour.

\begin{table}[htpb]
\centering
\rowcolors{2}{gray!25}{white}

\begin{tabular}{l|ccc}

\rowcolor{gray!40}

\textbf{Year} & \textbf{Precision} & \textbf{Recall} & \textbf{F1 Score} \\ \hline
$\mathbf{\ge2020}$ &     59.7      &   \textbf{ 60.3 }   &    \textbf{58.7}      \\ \hline
$\mathbf{\ge2018}$ &     59.6      &    58.0    &    57.9      \\ \hline
$\mathbf{\ge2015}$ &     61.1      &    56.0   &     53.9     \\ \hline
$\mathbf{\ge2010}$ &     63.4      &    55.6    &    52.8      \\ \hline
$\mathbf{\ge2005}$ &     \textbf{68.1}      &    56.5    &    52.0      \\ \hline
$\mathbf{\ge2000}$ &     66.1      &    56.8    &    51.8      \\ \hline
$\mathbf{\ge1990}$ &     65.6      &     55.4   &    51.3      \\ \hline
$\mathbf{\ge1980}$ &     64.2      &    54.7    &    50.0      \\ \hline

\end{tabular}

\caption{\label{tab:by_year} Results of final answer prediction over HealthFC-2, with different limitations on the earliest year of the retrieved evidence documents. All scores are macro averaged.}
\end{table}

\section{Results}
In this section, we present and describe the results of the conducted experiments. Table \ref{tab:topk} shows the final classification scores when changing the number of documents retrieved. Similarly, Table \ref{tab:topj} shows the final performance for different numbers of top sentences extracted. Tables \ref{tab:by_year} and \ref{tab:by_cits} show the influence of filtering evidence documents based on their year of publication and number of citations.

\subsection{Retrieved Documents and Sentences}
An interesting trend is observed when looking at Table \ref{tab:topk}. For all four datasets, the worst performance was when retrieving the highest amount of documents (50 and 100) and slowly increased towards the lower values. In the case of HealthFC-2 and TREC-Health, taking just the top document gave the best value of F1 (although the best macro precision and recall were with 5 and 10 documents). For BioASQ, taking into account just the top document is considerably worse than all other settings (because of poor recall), but the overall trend also holds for this dataset. As expected, the most challenging is the ternary version of HealthFC, but even there, the F1 performance increase of +2.5\% is observed from 100 to 1 document. For binary HealthFC and TREC-Health, jumps from 100 to 1 of +18\% and +10\% are seen.

When looking at Table \ref{tab:topj}, the results are less consistent than in the previous case. In fact, for datasets HealthFC-3 and BioASQ, the effect is rather opposite to the one in the previous experiment. The performance kept dropping as the number of selected sentences was lowered. The increased amount of knowledge in the bigger corpus of sentences helped the performance. For the binary HealthFC and TREC-Health, the numbers generally increased towards the lower number of sentences retrieved, but there wasn't a consistent pattern. Overall, the experiment showed that, in general, adding sentences that are relevant and semantically similar to the question increases the performance as opposed to adding more full documents to the QA system.

\begin{table}[htpb]
\centering
\rowcolors{2}{gray!25}{white}

\begin{tabular}{l|ccc}

\rowcolor{gray!40}
\textbf{\# Cits.} & \textbf{Precision} & \textbf{Recall} & \textbf{F1 Score} \\ \hline
$\mathbf{\ge100}$ &     59.7      &   \textbf{58.9 }   &    \textbf{59.1}      \\ \hline
$\mathbf{\ge75}$ &     59.6      &    57.9    &    58.0      \\ \hline
$\mathbf{\ge50}$ &     65.2      &    58.5   &     56.9     \\ \hline
$\mathbf{\ge25}$ &     65.0      &    57.8    &    55.8      \\ \hline
$\mathbf{\ge10}$ &     57.6   &    55.6   &    54.8      \\ \hline
$\mathbf{\ge5}$ &     \textbf{67.0}      &    56.9    &    53.6      \\ \hline
$\mathbf{\ge0}$ &     66.4      &    56.5    &    53.1      \\ \hline

\end{tabular}
\caption{\label{tab:by_cits} Results of final answer prediction over HealthFC-2, with different limitations on the minimum number of citations of the retrieved evidence documents. All scores are macro averaged.}

\end{table}

\begin{table}[pb]
\centering

\begin{tabular}{l|l|ccc}
 \multicolumn{2}{c}{} & \textbf{average} & \textbf{median} & \textbf{std.dev. } \\ \hline
\multirow{2}{*}{\shortstack[l]{Year}} & \tikzcmark &     2012.6     &   2014   &    8.1      \\ 
& \tikzxmark &     2009.3      &    2011    &    9.8      \\ \hline
\multirow{2}{*}{\shortstack[l]{\# Cits.}}  & \tikzcmark &     73.7      &    30    &    171.9     \\ 
& \tikzxmark &     63.6   &    26   &    122.2      \\ \hline

\end{tabular}
\caption{\label{tab:avgmedian} Average value, median, and standard deviation of years and citations of top 10 evidence documents retrieved for questions from HealthFC-3. They are grouped by those documents that yielded correct predictions (\tikzcmark) and those with incorrect predictions (\tikzxmark).}

\end{table}

\begin{table*}[!htpb]
\scriptsize
\centering
\begin{tabular}{p{20mm} p{41mm} p{41mm} p{41mm} }
\hline
\small \textbf{Question }      & \small  \textbf{Document \#1 }         & \small \textbf{ Document \#2 }  & \small \textbf{ Document \#3 }     \\ \hline
\raggedright \textbf{Is intense physical activity associated with longevity?} (\textbf{\textcolor{teal}{Supported}})                    &    Evidence, mainly from cross-sectional studies, suggests that physical activity is a potentially important modifiable factor associated with physical performance and strength in older age. It is unclear whether the benefits of physical activity accumulate across life or whether there are sensitive periods when physical activity is more influential. \cite{Cooper2011-vi} (\textbf{\textcolor{blue}{Not Enough Info}}) & Physical activity plays an important role for achieving healthy aging by promoting independence and increasing the quality of life. (...) Indeed, there is evidence to suggest that increasing exercise intensity in older adults may be associated with greater reductions in the risk of cardiovascular disease and mortality. \cite{El_Hajj_Boutros2019-mo} (\textbf{\textcolor{teal}{Supported}})       &  Exercise training above the public health recommendations provides additional benefits regarding disease protection and longevity. Endurance exercise, including high-intensity training to improve cardiorespiratory fitness promotes longevity and slows down aging. Strength training should be added to slow down loss of muscle mass, associated with aging and disease.   \cite{Pedersen2019-xd} (\textbf{\textcolor{teal}{Supported}})     \\
                     &          &          &                 \\
                     
\raggedright \textbf{Is dexamethasone recommended for treatment of intracerebral hemorrhage?} (\textbf{\textcolor{red}{Refuted}})                    &   Dexamethasone contributed to many serious adverse events. Conclusions: Given the small sample size, these preliminary results have not shown a clear beneficial effect of dexamethasone against placebo in our patients.  \cite{Prudhomme2016-qk} (\textbf{\textcolor{red}{Refuted}}) &  Overall, there is no evidence of a beneficial or adverse effect of corticosteroids in patients with either SAH or PICH. Confidence intervals are wide and include clinically significant effects in both directions. \cite{Feigin2005-vc} (\textbf{\textcolor{blue}{Not Enough Info}})       & Dexamethasone has been used to treat cerebral edema associated with brain abscess. (...) Conclusions: In patients with a brain abscess treated with antibiotics, the use of dexamethasone was not associated with increased mortality.  \cite{Simjian2018-hs}  (\textbf{\textcolor{teal}{Supported}})     \\

  \hline

\end{tabular}
\caption{\label{tab:top3} Example questions and retrieved evidence from the top three documents. In the first case, retrieving the top-3 performs better than retrieving just the top-1 document. In the second case, retrieving just the single best document gives a more precise verdict.}
\end{table*}

\subsection{Evidence Quality}

Table \ref{tab:by_year} tested the influence of the recency of the paper (year of publication) on the predictive performance. Once again, an interesting trend can be noted. The more recent the selected documents were, the more accurate the answers to health questions in our system. A similar phenomenon can be observed in Table \ref{tab:by_cits} for the number of citations of the papers. When limiting the selection to more and more cited papers, the final score kept increasing. Nevertheless, the highest scores in Tables 3 and 4 are still lower than the top-1-document performance from Table \ref{tab:by_year}. 

It intuitively makes sense that the more recent papers will provide the latest knowledge and insights into a research hypothesis, which then also better aligns with the gold labels of our datasets. This is also slightly biased by the recent creation date of our datasets, where annotators had access to the most recent knowledge. Likewise, the better the reputation of a paper (more citations), it could be assumed it will be a more reliable indicator of a correct answer. This wasn't the case for all of the questions, and there were many examples where older or less cited papers aligned better with the gold labels. Still, Table \ref{tab:avgmedian} provides some statistics that show that the trend generally holds. On average, those documents that produced a correct verdict were around three years more recent (in both the mean and median) while having an average of 10 citations more (median four citations). On the other hand, all categories have considerable standard deviations, showing many outliers and exceptions to this rule. 

Another challenging factor is that there could seemingly be an inverse correlation between the age of a paper and its number of citations -- older publications have had more time to gather a bigger number of citations. After deeper inspection, we observed that being cited a lot over time is only true for high-quality studies. Overall, striking a balance by finding both those studies that are both recent and already highly cited is an optimal strategy.

\begin{table*}[!htpb]
\small
\centering
\begin{tabular}{p{21mm} p{63mm} p{63mm} }
\hline
 \textbf{Question }      &   \textbf{Document \#1 }         &  \textbf{Document \#2 }     \\ \hline
\raggedright \textbf{Can stress promote dementia?} (\textbf{\textcolor{teal}{Supported}})                    &    To test the hypothesis that high job stress during working life might lead to an increased risk of dementia and Alzheimer's disease (AD) in late life. (...) Lifelong work-related psychosocial stress, characterized by low job control and high job strain, was associated with increased risk of dementia and AD in late life, independent of other known risk factors. \cite{Wang2012-ab} (\textbf{\textcolor{teal}{Supported}}) [\textbf{121 citations}]    & Patients with Alzheimer's disease (AD) or dementia are increasing in numbers as the population worldwide ages. Mid-life psychological stress, psychosocial stress and post-traumatic stress disorder have been shown to cause cognitive dysfunction. The mechanisms behind stress-induced AD or dementia are not known. \cite{Zhu2021-bq} (\textbf{\textcolor{red}{Refuted}}) [\textbf{3 citations}]             \\
                     &                    &                 \\

\raggedright \textbf{Can ginkgo biloba relieve the symptoms of tinnitus?} (\textbf{\textcolor{red}{Refuted}})                    &  Ginkgo biloba is a plant extract used to alleviate symptoms associated with cognitive deficits, e.g., decreased memory performance, lack of concentration, decreased alertness, tinnitus, and dizziness. Pharmacologic studies have shown that the therapeutic effect of ginkgo(...) \cite{Soholm1998-dq} %
(\textbf{\textcolor{teal}{Supported}}) [\textbf{Year:} \textbf{1998}] & We identified three systematic reviews including four primary studies, all corresponding to randomized trials. We concluded the use of Ginkgo biloba probably does not decrease the severity of tinnitus. In addition, it does not reduce the intensity of tinnitus or improve the quality of life of patients. \cite{Kramer2018-gj} (\textbf{\textcolor{red}{Refuted}})  [\textbf{Year: }\textbf{2018}]            \\
                
  \hline

\end{tabular}
\caption{\label{tab:year_cits} Example questions and retrieved evidence from two different documents, where only one of them provided a correct final answer. The more recent and the more cited papers provided better performance.}
\end{table*}

\section{Discussion}
In this section, we discuss and provide deeper insights uncovered in the results.

\subsection{Qualitative Analysis}
Specific effects of retrieving multiple documents for open-domain health question answering are shown in Table \ref{tab:top3}. For the first question, just retrieving the best document would have led to a study that does not provide a definitive conclusion to the question. However, the second and third most similar documents that were retrieved support the given research hypothesis. This held true even for further documents that cannot be shown in the table because of space constraints. On the other hand, the second question would have been more appropriately assessed by just looking at the best document instead of the top 3 documents. In fact, the second and third documents do not explicitly talk about the given question but rather a variation.

Examples of positive influences of taking into account the qualitative properties of the evidence into account, namely the year of publication and number of citations, are shown in Table \ref{tab:year_cits}. For the first question, a publication with above 100 citations directly answers the question and matches the gold annotation for this claim from the dataset. On the other hand, the second most similar retrieved document was a study with only three citations, which seems to be inconclusive about or even refuting the research hypothesis. Similarly, the answer to the second question on the effects of ginkgo is skewed by the top document from the 1990s that talks about the presumed positive effects of ginkgo. Two decades later, another meta-review analysis of systematic reviews on ginkgo biloba showed that it was not proven to help with tinnitus. This nicely demonstrates the changing nature of scientific knowledge and scientific consensus throughout time.

\subsection{Future Directions}

Based on our findings and discussion, we see that future work could focus on these directions:
\begin{itemize}
    \item  \textbf{Strength of evidence.} Taking into account the year of publication and number of citations has proven to be an effective strategy for enhancing the health question-answering performance. Similarly, further metadata could be taken into account to augment the process. In medical research, the strength of evidence is an important factor, and systems like GRADE are used to assess it \cite{balshem2011grade}. Different types of studies, such as a single study, a randomized clinical trial, and a systematic review, all have different strengths. Including this could be an important factor in improving the reliability of answers.
    \item  \textbf{Evidence disagreement and variation.} We observed how different studies and sources can come to differing conclusions regarding a claim. In this paper, we chose the majority vote among the \textit{top k} documents as the final decision, but this diminishes the information about the prediction uncertainty. This is part of the broader ML problem of learning with disagreements \cite{leonardelli-etal-2023-semeval} and modeling human label variation \cite{baan-etal-2024-interpreting}. While usually applied to uncertainty in data annotation, it could also be applied in the future to uncertainty in answering questions with diverse evidence documents.
     \item \textbf{Interpretability and user-centric results.} Other than just predicting the final answer, the end users posing biomedical and health questions would appreciate making the results more interpretable. This includes aspects such as displaying the different evidence documents, highlighting important sections, and showing prediction probabilities. Modern large language models (LLMs) could be used to enhance the reasoning process and to generate user-friendly explanations of model predictions and decisions \cite{10.1162/tacl_a_00398}.
\end{itemize}

\section{Conclusion}
In this paper, we conducted a number of experiments assessing the performance of a health question-answering system in an open-domain setting. Moving away from the standard setup of working with a small evidence corpus, we expand the knowledge sources to a large corpus of more than 20 million biomedical research abstracts. We measured the answer prediction performance over three diverse datasets of health questions while varying four aspects: number of retrieved documents, number of extracted sentences, and evidence quality in the form of year of publication and number of citations. Our results show that a lower number of documents retrieved leads to better performance, with the ideal spot in the 1--5 range. We also show that the performance is improved and made more reliable by using a time-aware evidence retrieval process, i.e., retrieving only the highly cited and more recent papers. Our research leaves room for exploration of disagreeing and conflicting evidence, generating explanations for end users, and including further metadata. We hope our research will encourage more exploration of the open-domain health question answering setting and addressing real-world user needs.

\section*{Limitations}

The question-answering pipeline used in this paper is a complex system with multiple factors -- the choice of the retrieval method, the sentence embedding model, the QA model, and the prediction threshold. It is possible that some incorrect predictions weren't due to the change in parameters but due to the faulty answer prediction. To account for this, we kept the reader part of the pipeline constant and fixed, to ensure a comparable setup. We focused on reporting only those findings and patterns that we observed were commonly occurring after a thorough analysis of retrieved evidence for each claim. Additionally, the final answer prediction for a given question was done by taking the label from the majority vote. This discards the information about evidence disagreement and label variation, which are also crucial in this domain. Future work should focus on this challenge.

\section*{Ethical Considerations}
The study focuses on the medical domain and answering health-related questions. This is a sensitive field where problems like misinformation, model hallucination, and incorrect evidence retrieval can lead to harmful consequences, disinformation spread, and societal effects. The question-answering system in this study shows promising performance but is still not ready for deployment and widespread use by end users, considering incorrect predictions, hallucinations, and lack of model interpretability.

\section*{Acknowledgements}
This research has been supported by the German Federal Ministry of Education and Research (BMBF) grant 01IS17049 Software Campus 2.0 (TU München).

\bibliography{anthology,custom}
\bibliographystyle{acl_natbib}

\appendix

\end{document}